\documentclass[draftcls,onecolumn,11pt]{IEEEtran}
\usepackage{algorithm}
\usepackage{algpseudocode}
\usepackage{listings}
\usepackage{balance}
\usepackage{xcolor}
\usepackage{soul}
\usepackage{graphicx}
\usepackage{amsmath,amstext}

\usepackage{amssymb}

\begin{document}


\title{Decentralised and collaborative machine learning framework for IoT}

\author{Martín González-Soto and Rebeca P. Díaz-Redondo and Manuel
  Fernández-Veiga and Bruno Rodríguez-Castro and Ana Fernández-Vilas\thanks{atlanTTic - I\&C Lab - Universidade de
    Vigo} \thanks{This is an extension of the conference paper "XuILVQ: A River Implementation of the Incremental Learning Vector Quantization for IoT", DOI: https://doi.org/10.1145/3551663.3558676.}}

\maketitle

\begin{abstract}
  Decentralised machine learning has recently been proposed as a potential
  solution to the security issues of the canonical federated learning
  approach. In this paper, we propose a decentralised and collaborative
  machine learning framework specially oriented to resource-constrained
  devices, usual in IoT deployments. With this aim we propose the following
  construction blocks. First, an incremental learning algorithm based on
  prototypes that was specifically implemented to work in low-performance
  computing elements. Second, two random-based protocols to exchange the local
  models among the computing elements in the network. Finally, two
  algorithmics approaches for prediction and prototype creation. This proposal
  was compared to a typical centralized incremental learning approach in terms
  of accuracy, training time and robustness with very promising results.  
\end{abstract}

\section{Introduction}
\label{sec:intro}
Decentralized machine learning faces how to use data and models from different
sources to build machine learning models that gather the partial knowledge
learned by each agent in this network to create, in a collaborative way, a
global vision or model of the whole network. This would allow processing large
amount of data managed by different computing elements. However, this approach
entails several issues that must be considered when proposing solutions for
this kind of computing environments. One of the most worrying is how to
provide secure and private solutions that protect personal data when building
global models.

Some approaches have been already proposed to decentralise machine learning
algorithms so that a set of networked agents can participate in building a
global model. Among them, the best known is federated learning
\cite{yang2019federated, zhang2021survey}. This approach is based on a
hierarchic organisation where a master computing element receives information
from other computing nodes in the network. The latter only share information
of the local model, but they do not share local data. The master element
combines these models to obtain a global one that is sent back to all
computing nodes. However, there have been arisen other
alternatives~\cite{hegedHus2019gossip, warnat2021swarm} that omit the central
server and try to provide a decentralised network, having the well-known
advantages of not having a central point of failure and potential attacks.

Additionally to these efforts to provide more robust and efficient
decentralised architectures for machine learning, there are aspects that must
be taken into account when selecting the algorithmics for these collaborative
scenarios. The philosophy of incremental learning~\cite{Gepperth2016,
  Ditzler2015, Losing2018} or online learning is a relevant approach that
refers to the continuous and adaptive learning of an algorithm. These
algorithms focus on creating models that can continuously learn as new
unobserved samples are introduced to the model from any data stream or
source. Thus, incremental learning allows a model to evolve and react to
perceived changes in its environment dynamically. However, it must be noticed
that this approach implies that the learning process must rely on more compact
representations, under memory limitation scenarios~\cite{Gepperth2016}.

The ability of a machine-learning algorithm to incrementally learn as new data
becomes available is a characteristic that allows machine-learning algorithms
to be more suitable for many current real-life problems such as data analysis
and big data processing in data streaming scenarios, robotics, or
decision-making systems. This is especially interesting for Internet Of Things
(IoT) applications because of two main reasons. First, in IoT solutions, data
is usually produced and extracted incrementally by massive sensor
networks. Second, IoT solutions often combine different kinds of devices
working cooperatively in hybrid Cloud-Fog-Mist architectures
\cite{ketu2021cloud}, where low-performance devices also participate in the
Fog-Mist layer. Therefore, incremental learning algorithms allow
low-performance sensors/actuators to participate in the learning process of
dynamic data streams.

Within this context, IoT solutions with low-performance devices processing
data streams, we propose a collaborative architecture were computing nodes
work under a decentralised policy to jointly create a global model by applying
an incremental learning approach. Being more specific, our contributions are
the following ones:
\begin{itemize}
\item We have adapted an incremental learning algorithm, ILVQ (Incremental
  Learning Vector Quantization)~\cite{Xu2011}, to properly work with
  resource-constrained devices, typical ones in IoT deployments. We have
  implemented this modification of the algorithm, coined as XuILVQ, and made
  it publicly accessible in the Python River online learning
  library~\cite{harries1999splice}. Contributions made to the algorithm
  include adding a mechanism for class prediction and rearranging the
  calculation of the threshold distance to consider all initial cases.
  
\item Therefore, our XuILVQ implementation is the building block of a
  decentralised and collaborative network of devices. We propose two different
  incremental learning mechanisms. One the one hand, a Fully ILVQ approach,
  where both predictive model generation and prototype generation is performed
  by the XuILVQ algorithm. On another hand, a Hybrid approach, where the
  XuILVQ algorithm is in charge of generating the prototypes to share with the
  other nodes in the network and another incremental algorithm, the Adaptive
  Random Forest (ARF) is in charge of the predictive tasks.
  
\item In order to provide a decentralised collaborative learning framework, we
  have defined two protocols for the nodes to share their prototypes with
  their peers. The first one, the random sharing protocol does not take into
  account the performance context: neither the local performance nor the
  global performance. The prototypes are randomly shared with a subset of peer
  nodes. The second one, the relative threshold protocol share its local model
  to other nodes only when these ones are underperforming. By utilizing
  prototypes as pieces of information to be shared, these protocols become
  more interpretable for a user as they are synthetic samples that capture the
  observations of the model in a synthesized form. Conversely, since the model
  introduces the prototypes as samples to its model, they can be introduced or
  omitted as new prototypes depending on their ability to represent previously
  unobserved cases, meaning on the information contained in them. This differs
  from other proposed approaches in which the model parameters are only shared
  without any mechanism to distinguish valuable information from non-valuable
  information.
  
\item Finally, we have conducted an evaluation of the performance of the
  proposed system (with all the previous variants) according to their
  accuracy, training time and robustness. This evaluation is done by comparing
  our approach to the traditional centralized incremental learning system.
\end{itemize}

The rest of this paper is organized as follows. Section~\ref{sec:background}
summarizes previous approaches in the literature for prototype-based learning
and incremental adaptations of these algorithms. Section~\ref{sec:proposal}
presents the problem and the proposal on which this paper
focuses. Section~\ref{sec:system_model} describes the material and methods
needed to carry out the proposed experiment. Section~\ref{sec:results}
presents the main components and the structure of the implementation of ILVQ
in River.  Finally, conclusions will be presented in
Section~\ref{sec:conclusions}.

\section{Background}
\label{sec:background}

This Section presents and develops the fundamental concepts discussed in the
article.  First, the concept of prototype-based learning is introduced,
starting from a general perspective and leading to specific
details. Subsequently, we introduce the Incremental Learning Vector
Quantization\cite{Xu2011} model, a fundamental element of our
contribution. The section also examines collaborative learning, analyzing
proposed techniques that form the foundation for our framework.

\subsection{Prototype-based learning}

Prototype-based learning~\cite{Biehl2016} is a family of techniques in the
field of machine learning that makes use of two concepts. (1) Explicit
representation of observations through representative examples, i.e.,
prototypes; (2) Comparison between observations and stored prototypical
examples using a similarity function, typically a distance metric.

Within the family of prototype-based learning algorithms, there are techniques
for unsupervised learning such as Competitive Vector Quantization
(VQ)~\cite{AHALT1990277}, Neural Gas Algorithm
(NG)~\cite{martinetz1991neural}, or Self-Organizing Maps
(SOM)~\cite{kohonen1982self}. These techniques can be useful to reduce the
number of stored samples or find clusters. Among the supervised machine
learning techniques, $k$-nearest neighbors (kNN)~\cite{fix1951discriminatory}
is one of the most popular and simplest algorithms. Learning Vector
Quantization (LVQ)~\cite{kohonen1995} is an alternative to kNN that keeps a
smaller number of elements in memory by keeping a representative set of
prototypes instead of all the samples in memory.  As a result, LVQ is a
flexible, easily interpretable supervised algorithm with competitive
performance in some practical applications.

\subsection{Learning Vector Quantization}

Learning Vector Quantization is a supervised prototype-based learning algorithm proposed by
Kohonen~\cite{kohonen1995}. Its behavior is based on an update of the prototypes vectors when a feature 
vector $\mathbf{x}_u \in \mathbb{R}^{n}$  and a class label $y_u \in \mathbb{R}$ are sampled. The closest 
prototype (in Euclidean distance) $\mathbf{q}^\ast = Q(\mathbf{x}_u) := \inf_{\mathbf{q} \in \mathcal{P}} 
\| \mathbf{x}_u - \mathbf{q} \|$ to the new sample $\mathbf{x}_u$ is chosen 
from a dictionary $\mathcal{P}$, and it is updated based on the match between prototype and sample class, the distance 
between them, and learning rate $\eta$
\begin{equation}
  \label{eqn:updaterule}
  \mathbf{q}^\ast \leftarrow \mathbf{q}^\ast + \eta \psi(\hat{y}, y_u)
  (\mathbf{x}_u - \mathbf{q}^\ast), 
\end{equation}
where $\hat{y}$ is the predicted class (the class of the prototype), and
\begin{equation}
  \psi(\hat{y}, y) = 
  \left\{
    \begin{array}{ll}
      +1  & \text{if $\hat{y} = y$} \\
      -1 & \text{otherwise}.
    \end{array}
  \right.
\end{equation}
A similar adaptive mechanism for updating the prototype feature vector is
shared by a number of variants of this algorithm, and also by the algorithm we
propose in this paper.

Many modifications have been proposed to improve LVQ convergence speed,
generalization capacity, or approaches based on cost functions. However, in
this paper, we only focus on the proposed modifications to transform it into
an incremental algorithm.

In~\cite{Losing2015}, it is proposed the use of four mechanisms to adapt the
LVQ algorithm to incremental behavior: (1) A short-term memory $\psi$
containing the more recent $t$ samples; this short-term memory is used to
sample new prototype candidates when needed; (2) An inclusion mechanism based
on error counting; (3) A prototype insertion policy that includes the
candidate proposed by the placement strategy and updates the distances in
$\psi$; (4) And a placement strategy that selects the candidate from $\psi$
that minimizes the cost function proposed by Sato and
Yamada~\cite{sato1995generalized}.

Chen and Lee \cite{chen2020incremental} propose an incremental few-shot
learning algorithm that makes use of deep incremental learning vector
quantization to solve problems related to catastrophic forgetting in
class-incremental tasks. A self-incremental learning vector quantization
algorithm (SILVQ) is proposed in~\cite{manome2021self}; this algorithm is
capable of learning concepts while adjusting the learning rate by
incorporating symmetric bias and mutually exclusive bias into learning vector
quantization. In~\cite{shen2020online} it is proposed a method for online
semi-supervised learning based on the use of prototype-based learning. The
proposed algorithm combines the use of learning vector quantization for
labeled data and the use of the Gaussian mixture clustering criterion / neural
gas clustering criterion to obtain valuable knowledge from the unlabeled data
that the algorithm receives.

\subsection{Incremental Learning Vector Quantization}

Incremental Learning Vector Quantization is a modification of the
prototype-based LVQ algorithms~\cite{kohonen1995} that aims to build an
incremental prototype-based model proposed by Xu et al.~\cite{Xu2011}. This
adaptation is made through three contributions: (1) The first contribution is
designing a mechanism to include new prototypes and ensure incremental
between-class learning and incremental within-class learning. On the one hand,
it seeks to introduce samples of classes that have not been observed during
the training process. On the other hand, for within-class learning, a sample
will be added to the set of prototypes only when this sample, being of the
same class, is sufficiently different from those already contained in the set
of prototypes (threshold trespassing condition). (2) The proposed threshold
for the prototype inclusion process must be adaptive so that the set of
prototypes does not increase indefinitely over time. Therefore, the authors
propose an adaptive threshold that will keep a reduced number of prototypes to
represent each class in memory. (3) The last contribution of the algorithm is
a mechanism for eliminating obsolete prototypes. The mechanism penalizes older
prototypes that have not been recently employed as the closest prototype to a
training sample.

During its operation, the algorithm may receive a previously unobserved class,
in which case the algorithm will register the sample as a prototype that
represents the class and will therefore have information about that class in
the next iteration and it will no longer be unknown.

\subsection{Collaborative Learning models}

In a collaborative learning environment, the objective is to train a machine
learning model in a cooperative way by a set of computation nodes that receive
local samples. Each computation node only learns from its own data but, in
order to improve the global performance, these nodes share or exchange
relevant information to refine their own model. Therefore, each one learns
from the local data set and also from the knowledge acquired by its peers
(other nodes).

One of the techniques that recently arose for this kind of scenarios is the
federated learning approach \cite{yang2019federated,
  zhang2021survey}. Federated learning provides a framework designed to
protect user privacy, which makes it ideal for applications where sensitive
information is handled. Its canonical architecture is a hierarchy where a
central computation element receives information from each dependent node,
then combines this knowledge to create a better learning model, which is later
shared back with all the computation nodes. The information shared is the
learning model parameters, usually encrypted, with the aim of protecting the
data, which is not shared.

The typical workflow involves a set of $M$ nodes in the network
$\{N_1, ... , N_m\}$ with their own local databases $\{D_1, ..., D_m\}$ that
cannot be accessed from any other node in the network. Usually (1) the central
server assigns each node an initial model; then (2) each node $N_i$ locally
trains its own model $W_i$ with its local data set $D_i$; after that, (3) the
central server aggregates the local models obtained from each node
$\{W_1, ..., W_n\}$ to create a global model $W'$; finally (4) this global
model is sent back to each node to repalce the local model.

Despite its widespread use and numerous applications, federated learning
implies a total dependence on a central server to operate correctly. This
entails risks, such as susceptibility to failures and vulnerability to
attacks. Therefore, other alternatives are emerging to try to offer a
decentralized solution.

In \cite{hegedHus2019gossip} it is proposed a fully decentralized alternative,
coined as Gossip learning, to federated learning based on the Gossip
algorithm. This method assumes a set of nodes $\{N_1, ..., N_m\}$ that have a
local model $W_i$ and its time stamp or age $t_i$. This information is
periodically shared with other nodes in the network. When a node receives the
model of one of its peers, it combines both its local model and the one that
has been received to create an updated local model. This combination is
typically done by averaging the parameters of both models and, finally, the
old local model is overwritten by the combined one. This new local model
(combined) continues being trained with the local data $D_i$.

The Swarm Learning (SL) model \cite{warnat2021swarm} is another alternative to
the centralized philosophy. In this approach, each local model $W_i$ is
trained with local data, as in the Gossip learning model, but in this case, an
extra step is used to guarantee the traceability and, in consequence, to
increment the security of the whole system. The blockchain technology
\cite{liu2020blockchain} is used to share the local models (parameters) among
the peer nodes in the network. Each new node subscribes through a smart
contract, obtains the model and performs local training until the requirements
defined in the smart contract are met. The model parameters are then exchanged
and combined to create an updated model.

\section{Problem description and proposed system}
\label{sec:proposal}

We consider a network of $n$ nodes running independent instances $M_i$,
$i = 1, \dots, n$, of an incremental learning model $\mathcal{M}$. Each node
has access to a local data set $\mathcal{D}_i$ for training and inference,
these data sets are independent and identically distributed. The incremental
learning algorithms that we propose, prototype-based models, are described in
Section~\ref{sec:ml_algorithms}.

Any two nodes can communicate directly through wireless links, and for
simplicity it is assumed that transmissions are free from errors. Each node
can be thought of as a low-computational device equipped with sensors to
collect data sequentially. The communication among them will be used to
exchange relevant information (the local models $M_i$, no data) and build an
improved global model. The exchange of local models (prototypes) is done by
using decentralised protocols, which are described in
Section~\ref{sec:protocols}.

\begin{figure}[t]
  \centering \includegraphics[width=\columnwidth]{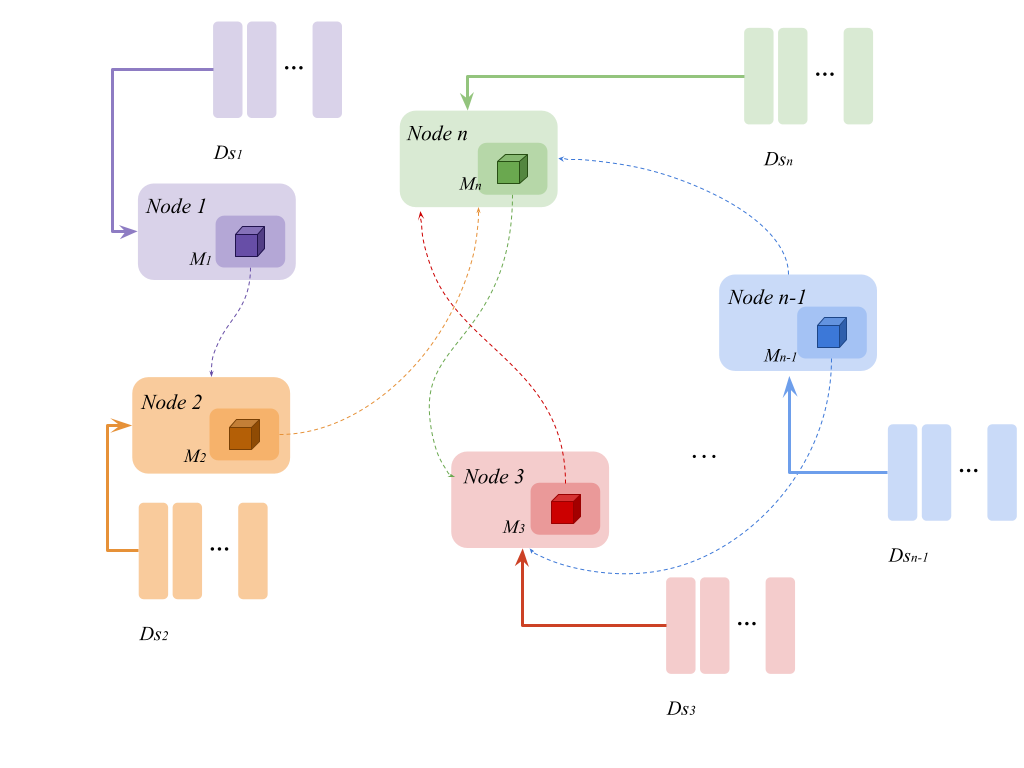}
  \caption{Example configuration of nodes within a network of devices sharing
    information. Each node represents a physical device with low computational
    capacity, equipped with sensors to collect data that is fed into the
    incremental learning model. }
  \label{fig:architecture-general}
\end{figure}

Therefore, our approach to improve the performance of network models, shown in
Figure~\ref{fig:architecture-general}, involves leveraging sharing protocols
in combination with prototype-based models. The decentralised protocols enable
the distribution of knowledge learned locally (individual models $M_i$) to the
whole network, thereby increasing the diversity and quantity of available
data. The generated prototypes act as a form of learned knowledge,
encapsulating both synthetic samples summarizing the context of a node and the
associated model parameters. The dual nature of the prototypes makes them a
suitable means of synthesizing the context of a node, providing a concise and
informative representation of the knowledge to be transferred to another node
for learning purposes.

By leveraging sharing protocols and prototype-based models, it becomes
feasible to transfer knowledge between nodes and attain a collection of global
models in a decentralised manner. In addition, because the algorithms within
each device are incremental, the framework maintains its incremental quality
over time.

\subsection{Decentralised Prototype Sharing Protocols}
\label{sec:protocols}

Nodes in the collaborative learning network share their local models with the
other nodes to enhance the model performance. With this purpose, we have
defined two different decentralised prototype sharing protocols. Both of them
are used to assess if the performance under this scheme improves the
performance of the traditional centralised approach. We have included a
section where the design of the algorithms is mathematically justified in the
Appendix~\ref{secA1}.

On the one hand, we propose a random sharing protocol, when each node randomly
decides when to share the local model and with which other nodes this
information is shared. On another hand, we propose a more refined protocol,
coined as relative threshold protocol, where the model is shared only if the
node is outperforming compared to the other nodes in the network. The notation
we use to detail both protocols is summarized in
Table~\ref{tab:relativeprotocol}.

\begin{table}[t]
  \centering
  \caption{Decentralised sharing protocols: Notation. Parameter $C_i$ only
    applies in the relative threshold protocol }
    \label{tab:relativeprotocol}
    \begin{tabular}{c c} \hline
         Notation & Description \\
         \hline
         $M_i$ & Local model in node $i$\\
         $G_i$ & Prototype set of $M_i$\\
         $Q_i$ & Queue of received prototypes of $M_i$\\
         $t$ & Sharing threshold\\
         $s$ & Relative subset size\\
         $C_i$ & Hash table with performance values of other nodes in node $i$ \\ 
         \hline\\
    \end{tabular}
\end{table}

\paragraph{Random sharing protocol} This protocol, whose pseudo-code is
detailed in Algorithm~\ref{alg:rsp}, is a basic method that does not take into
account the performance context: neither the local performance nor the global
performance. Thus, at each training iteration (new sample) each node randomly
decides to share the local model (prototypes) or not. If so, local prototypes
($G_i$) are sent to a random subset of the other nodes in the network. This
protocol needs two parameters: $t$, the probability of sharing the local model
at each iteration, and $s$, the number of nodes in the network with which the
local model is going to be shared. Each node has an associated queue where
other nodes can write their prototypes for sharing. After processing a new
sample the node will share or not its set of prototypes based on the $t$
parameter. In case it is going to share, it will write its prototypes in the
queues of each sampled node. Finally, it will process all prototypes in its
queue before moving to the next sample.

\begin{algorithm}[t]
  \caption{\label{alg:rsp}  Random sharing protocol}
  \begin{algorithmic}[1]
    \State Initialize $M_i$
     \State Input new pattern $\mathbf{x} \in \mathbb{R}^n$ (with label $y \in \mathbb{R}$)
        \If{$K \sim U(0, 1) < t$}
            \State Sample a subset of nodes of size $s$ in the network
            \State Send $G_i$ to the subset of nodes
        \EndIf
    \While{$Q_i  \neq \emptyset$} \Comment{Process all the received prototypes before the next sample}
        \State Input prototype $\mathbf{x} \in \mathbb{R}^n$ (with label $y \in \mathbb{R}$)
    \EndWhile
    \State Go to step 2 to process a new input pattern
    \end{algorithmic}
\end{algorithm}

\paragraph{Relative threshold protocol} This protocol, whose pseudo-code is
detailed in Algorithm~\ref{alg:rtp}, shares its local model to other nodes
only when these other nodes are underperforming. This protocol also needs two
parameters: $t$, the probability of sharing the local model at each iteration,
and $s$, the number of nodes in the network with the local model is going to
be shared. However, it also needs to keep the prototypes (local models) $Q_i $
of the other nodes in the network and to calculate and record the performance
of the other nodes, besides of its own performance. Thus, with probability $t$
the local model is shared with a maximum number of nodes $s$ only if the
performance of these randomly selected nodes is lower than the local
one. Consequently, the local model will be shared only with a subset of nodes
that have low predictive ability. In a similar vein to the earlier protocol,
the prototypes shall be written to the queues of sampled
prototypes. Subsequently, the received prototypes in the queue shall be
processed prior to handling the next sample.

\begin{algorithm}[t]
  \caption{\label{alg:rtp} Relative threshold protocol}
  \begin{algorithmic}[1]
    \State Initialize $M_i$ and $C_i$
     \State Input new pattern $\mathbf{x} \in \mathbb{R}^n$ (with label $y \in \mathbb{R}$)
        \If{$K \sim U(0, 1) < t$}
            \State Sample a subset of nodes of size $s$ in the network
            \State Compute the median of the performance values of the sampled nodes
            \If{$C_i[i] > \operatorname{median}(C_i)$ }
                \State  Send $G_i$ to the subset of nodes who perform below the median
            \EndIf
        \EndIf
    \While{$Q_i  \neq \emptyset$} \Comment{Process all the received prototypes before the next sample}
        \State Input prototype $\mathbf{x} \in \mathbb{R}^n$ (with label $y \in \mathbb{R}$)
    \EndWhile
    \State Go to step 2 to process a new input pattern
    \end{algorithmic}
\end{algorithm}

\subsection{Machine learning algorithms}
\label{sec:ml_algorithms}

As it was previously mentioned, we propose to use an incremental algorithm
based on prototypes with the aim of having local models (prototypes) that are
also be considered as summary samples or enriched samples, since they
summarized the knowledge obtained during the algorithm execution.

We have decided to use as base the Incremental Learning Vector Quantization
(Algorithm~\ref{alg:ilvq}) proposed by Xu et al. \cite{Xu2011}, explained in
Section~\ref{sec:background}. This algorithm produces descriptive synthetic
samples (prototypes) as its model parameters, which are highly effective in
synthesising the learning context of a node and, consequently, they are ideal
to be shared with other nodes in the collaborative network.

Within this context, we propose to use the Incremental Learning Vector
Quantization (ILVQ) \cite{Xu2011}. In our previous work
\cite{gonzalez2022xuilvq}, from which this paper is an extension, we
introduced our modifications and implementation, which is detailed in
Algorithm \ref{alg:ilvq}. This new algorithm, coined as XuILVQ, was adapted
according to two aspects. On the one hand, to the Python's River incremental
learning library, to provide a easy way for other researchers to apply it. On
the other hand, to adapt the algorithm to devices with limited computational
capacity. The underlying idea is these devices were able to generate
incremental models based on the device's measurements, a typical scenario in
edge computing and IoT. With this aim, we have designed some modifications to
adapt the proposed model \cite{gonzalez2022xuilvq}.

Firstly, the threshold computation function
(Algorithm~\ref{alg:threshold_computation}) was adjusted to account for
potential initialization cases. The threshold computation determines whether a
sample is sufficiently distinct from previously observed samples, in which
case the sample is introduced as a new prototype to the model. This
computation is based on the prototypes' neighbors, and it is necessary to
consider situations where a prototype (i) lacks neighbors, (ii) lacks
neighbors of the same class, or (iii) lacks neighbors of a different class.

Secondly, we have added a prediction mechanism
(Algorithm~\ref{alg:ilvq-predict}) to the ILVQ algorithm implementation. Since
it is a prototype-based algorithm, we have opted for the nearest-neighbor
approach. Thus, using a defined parameter $k$, the algorithm identifies the
$k$ closest prototypes to a given sample, and the class will be assigned based
on the most frequent class among its neighbors.

\begin{algorithm}[p]
  \caption{\label{alg:ilvq} Incremental learning vector quantization}
  \begin{algorithmic}[1]
   \State Initialize $G$ to contain the first two input data from the
   training set
   \State Initialize $E$ storing connection between prototypes to
   empty set
   \State Input new pattern $\mathbf{x} \in \mathbb{R}^n$ (with label $y = \phi(\mathbf{x}) \in \mathbb{R}$) \label{alg:inputilvq}
   \State Search set $G$ to find the winner and runner-up by $\mathbf{s}_1 =
   \arg\min_{\mathbf{c} \in G} || \mathbf{x} - \mathbf{w_c} ||$  and
   $\mathbf{s}_2 = \arg\min_{\mathbf{c} \in G\setminus \{ \mathbf{s}_1 \}} || \mathbf{x} -
   \mathbf{w_c} ||$. \label{alg:closest}
   \If {($\mathbf{x}$ is in a new class) $\lor \| \mathbf{x} - \mathbf{s_1} \| > T_{\phi(\mathbf{s}_1)} \lor \| \mathbf{x} - \mathbf{s_2} \| > T_{\phi(\mathbf{s}_2)}$} \label{alg:threshold}
     \State $G \leftarrow G \cup \{ \mathbf{x} \}$ \label{alg:append}
     \State Go to step~\ref{alg:inputilvq} to process a new input pattern
    \EndIf
    \If {$(\mathbf{s}_1, \mathbf{s}_2) \not\in E$}
      \State $E \leftarrow E \cup \{ (\mathbf{s}_1, \mathbf{s}_2) \}$ \label{alg:create_edge}
      \State $\mathsf{age}(\mathbf{s}_1, \mathbf{s}_2) \leftarrow 0$
    \EndIf
%
        \For {$\mathbf{s}_i: (\mathbf{s}_1, \mathbf{s}_i) \in E$}
    \Comment{Iterates over the neighbor prototypes \label{alg:update_age_1}}
      \State $\mathsf{age}(\mathbf{s}_1, \mathbf{s}_i) \leftarrow \mathsf{age}(\mathbf{s}_1, \mathbf{s}_i)
      + 1$
    \EndFor\label{alg:update_age_2}
    \State $M_c \leftarrow M_c + 1$ \Comment{$M_c$ counts the number of points in class $c$} \label{alg:updatem}
        \If {$\phi(\mathbf{s}_1) = \phi(\mathbf{x})$} \label{alg:update_vector_1}
        \State {$\mathbf{w}_{s_1} \leftarrow \mathbf{w}_{s_1} + \mathbf{\eta_1} (\mathbf{x} - \mathbf{w}_{s_1})$} \Comment{Gradient update towards $\mathbf{x}$}
        \For{$\mathbf{s_i}: (\mathbf{s}_1, \mathbf{s}_i) \in E$, and $\phi(\mathbf{s}_i) \neq \phi(\mathbf{x})$}
            \State {$\mathbf{w}_{s_i} \leftarrow \mathbf{w}_{s_i} - \mathbf{\eta_2} (\mathbf{x} - \mathbf{w}_{s_i})$}
        \EndFor
    
    \Else \Comment{$\mathbf{x}$ is not in the same class as $\mathbf{s}_1$}
        \State {$\mathbf{w}_{s_1} \leftarrow \mathbf{w}_{s_1} - \mathbf{\eta_1} (\mathbf{x} - \mathbf{w}_{s_1})$} 
        \For{$\mathbf{s_i}: (\mathbf{s}_1, \mathbf{s}_i) \in E$ and $\phi(\mathbf{s}_i) = \phi(\mathbf{x})$}
            \State {$\mathbf{w}_{s_i} \leftarrow \mathbf{w}_{s_i} + \mathbf{\eta_2} (\mathbf{x} - \mathbf{w}_{s_i})$}
        \EndFor
    \EndIf \label{alg:update_vector_2}
    \State $E \leftarrow \{ (\mathbf{s}_1, \mathbf{s}_2) \in E\ |\ \mathsf{age}(\mathbf{s}_1, \mathbf{s}_2) < AgeOld \}$ \label{alg:edge_elimination} \Comment{prune old edges in $E$}
    \If{the iteration step index is the integer multiple of parameter $\lambda$} \label{alg:denoise_1}
        \State Delete the nodes $\mathbf{s}_i$ in the set $G$ that have no neighbor node.
        \State Delete the nodes $\mathbf{s}_i$ whose neighbor node is 1 and $M_{s_i} < 0.5 \sum_{j=1}^{N_G} \frac{M_j}{N_G}$ \Comment{Remove points in low-density class}
    \EndIf \label{alg:denoise_2}
    \State Go to step~\ref{alg:inputilvq} to process a new input pattern \label{alg:end}
  \end{algorithmic}
\end{algorithm}

\begin{algorithm}[t]
  \caption{\label{alg:threshold_computation} Modification of the distance threshold $T_i$ computation}
  \begin{algorithmic}[1]
  
  \State $n_i = \sum_{\mathbf{w}} 1 \bigl( \phi(\mathbf{w}) = i \bigl)$ \Comment {count the points in class $i$, $\phi(\cdot)$ gives the label}
  \State $\delta_i = \frac{1}{n_i} \sum_{(i, j) \in E: \phi(\mathbf{w}_i) = \phi(\mathbf{w}_j)}{|| \mathbf{w}_i - \mathbf{w}_j||}$ \Comment{Mean distance, in-class}
  \State $T_i = \min_{(i, j) \in E: \phi(\mathbf{w}_i) \neq \phi(\mathbf{w}_j)} || \mathbf{w}_i - \mathbf{w}_j||$ \label{alg:search_t} \Comment{Min distance, between-class} 
  \State $j^\ast = \arg \min \{ \| \mathbf{w}_i - \mathbf{w}_j \|: (i,j) \in E, \phi(\mathbf{w}_i) \neq \phi(\mathbf{w}_j \} $
  \If{$ E = \emptyset$}
  \State return none
  \EndIf
  \If{$\{ (i, j) \in E: \phi(\mathbf{w}_i) = \phi(\mathbf{w}_j)\} = \emptyset$}
  \State return $\max(\{ \| \mathbf{w}_i - \mathbf{w}_j \|: (i, j) \in E, \phi(\mathbf{w}_i) \neq \phi(\mathbf{w}_j) \})$
  \EndIf

    \If{$\{ (i, j) \in E: \phi(\mathbf{w}_i) \neq \phi(\mathbf{w}_j) \} = \emptyset$}
  \State return $\min(\{ \| \mathbf{w}_i - \mathbf{w}_j \|: (i, j) \in E, \phi(\mathbf{w}_i) = \phi(\mathbf{w}_j) \}$
  \EndIf
  
  \If{$T_i < \delta_i$}
  \State $T_i = \min \{ \| \mathbf{w}_i - \mathbf{w}_k \|: (i, k) \in E, \phi(\mathbf{w}_i) \neq \phi(\mathbf{w}_k), k \neq  j^\ast \}$
  \EndIf
  \State Go to step~\ref{alg:search_t} until $T_i \ge \delta_i$
  \State return $T_i$
  \end{algorithmic}
\end{algorithm}

\begin{algorithm}[t]
  \caption{\label{alg:ilvq-predict} XuILVQ predict\_one($\mathbf{x}$)}
  \begin{algorithmic}[1]
  \State $K \leftarrow$ $k$-nearest prototypes to the given feature vector $\mathbf{x}$
  \State $\mathbf{\hat{v}} \leftarrow \mathbf{0}$
  \For{$\mathbf{w_i} \in K$}
  \State $c \leftarrow \phi(\mathbf{w}_i)$ \Comment $\phi(\cdot)$ gives the label/class of a point
  \If{$||\mathbf{x}-\mathbf{w}_i|| = 0$}
        \State $\mathbf{\hat{v}}_c = 1 $
        \State Go to step~\ref{alg:return}
    \Else
        \State $\mathbf{\hat{v}}_c \leftarrow \mathbf{\hat{v}}_c + \frac{1}{||\mathbf{x} - \mathbf{w}_i ||}$
    \EndIf
  \EndFor
  \State  $\mathbf{\hat{v} \leftarrow Softmax(\hat{v})}$
  \State return $\arg\max_i \mathbf{\hat{v}}_i$ \label{alg:return}
  \end{algorithmic}
\end{algorithm}

The decentralised learning system that we propose works under two different
scenarios, according to the algorithm used for the predictive task.

\paragraph{Fully ILVQ approach} In this scenario we applied ILVQ for both
predictive model generation and prototype generation. Thus, upon arrival of a
new sample, the ILVQ algorithm processes this sample to update the predictive
model and also to generate a new prototype or to update the set of prototypes.

\paragraph{Hybrid approach} In this scenario we use a combination of ILVQ and
the Adaptive Random Forest (ARF) algorithm to separate the tasks of predictive
and prototype generation. Thus, the ARF algorithm processes new samples to
update the predictive model whereas ILVQ generates or updates the set of
prototypes. The prototypes can be shared among nodes as samples, but in this
case, they serve as inputs to the ARF algorithm instead of the ILVQ one.

\section{Experimental design}
\label{sec:system_model}

We now describe the experimental design and methodology that was employed to
test the proposed scheme.  The experiments were devised to evaluate the
potential use of the learning algorithm for generating prototypes that could
be used to transfer knowledge between models and enhance the diversity of
their local prototypes. To achieve this goal, we implemented the two
gossip-based prototype exchange protocols.  For assessing the ability of the
gossip algorithm in improving the quality of the global model and speed up its
convergence, we use a complete network of $5$ and $7$ nodes where each node
executes locally an instance of ILVQ on its local dataset. All the datasets
have the same distribution and are independent, and there is no overlap
between the different partitions. Asymmetry is introduced in this setting just
by using a partition in which one of the nodes works on a smaller part of the
whole dataset. Performance metrics for this tagged node are introduced below,
and the results will be reported and discussed in Section~\ref{sec:results}.

\subsection{Setup and performance metrics}

As our baseline case for comparison, we take the achievable performance in a
centralized scheme where a single node has access to the entire data stream
and runs ILVQ continuously. Our performance metrics are the following: (1)
F-score achieved by the consensus model,
$F = 2 \mathsf{tp} / (2 \mathsf{tp} + \mathsf{fp} + \mathsf{fn})$, where
$\mathsf{tp}$ is the number of true positives, $\mathsf{fp}$ is the number of
false positives, and $\mathsf{fn}$ is the number of false negatives; (2)
convergence time $T_c$ , defined as the time necessary to reach 85\% of the
best F-score; (3) storage complexity $M$: amount of memory dedicated to store
the dictionary of prototypes at each node; (4) communication complexity $L$:
the number of messages exchanged among the nodes with the gossip algorithm.

The two learning protocols presented in the previous section will be
evaluated, namely the pure ILVQ algorithm and the combined ILVQ+ARF approach
for prototyping and prediction tasks.  To assess the impact of the sharing
protocol, we will track changes in the F-score as the $T$ is varied.  While we
recognize that the parameter $s$ might introduce noise in our results due to
its potential absence in a significant proportion of samples, we have set it
to a value to be broadcast each time an update message is sent. The same
approach of simulating ILVQ and ILVQ+ARF will be followed with the
contextualized gossip protocol.

\begin{figure}[t]
    \centering
    \includegraphics[width=\columnwidth]{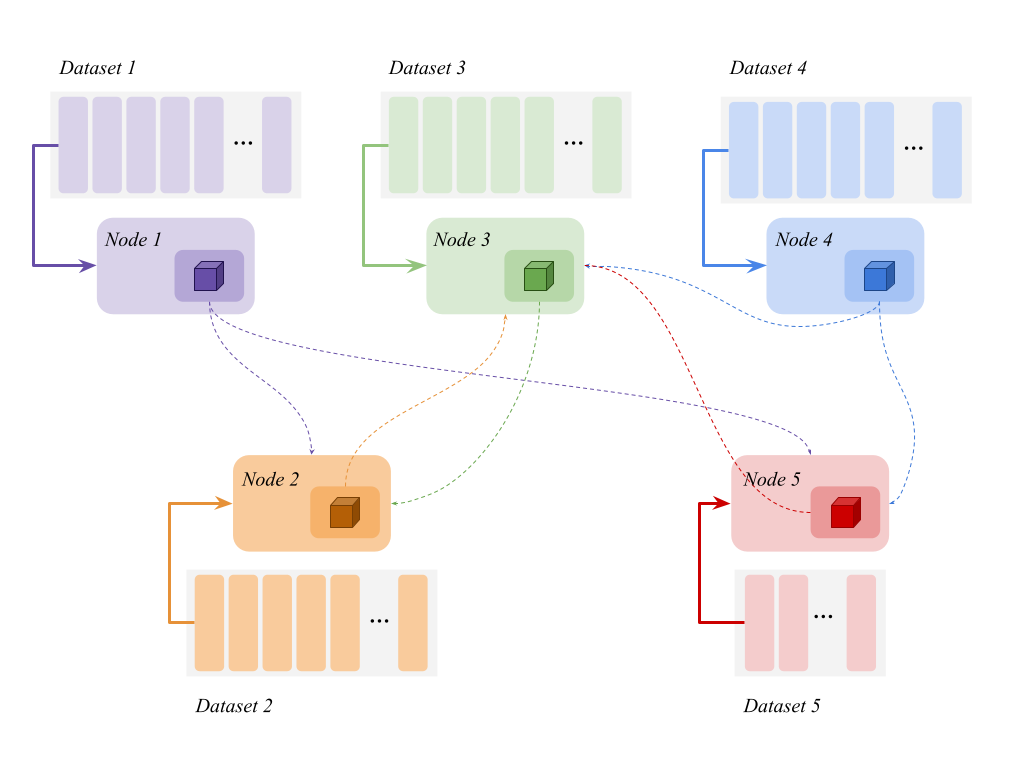}
    \caption{\label{fig:architecture} Architecture of the decentralised
      system: five nodes with different numbers of samples. Asymmetry in the
      size of the local datasets is generally assumed.}
\end{figure}

Our goal is to test a network of interconnected, low-performance devices and
simulate data arrival to these devices in a production scenario from various
sources, including sensor measurements. Each device is associated to an
incremental embedded model (ILVQ) that is trained with the data collected or
received by the device. Our objective is to evaluate the potential of
knowledge sharing between models to improve predictive performance. The choice
of ILVQ is dictated by the presumed application case, where a set of sensor
nodes actively or passively collect measurements from their neighborhood and
update a discriminative model. In turn, for the construction of the consensus
global model, we wish to minimize the use of communication resources, so
instead of exchanging the samples our protocols send the updated prototypes to
the other nodes. Thus, the prototypes and their labels summarize the dynamic
behavior of the classifiers.

\subsection{Dataset and partition} 

For our experiments, we use the \textsf{Phishing} dataset, which is part of
the Python River online learning
library~\cite{harries1999splice}. \textsf{Phishing} comprises $1250$ samples
and includes ten features that relate to the properties of a given set of web
pages. The objective is to utilize these features to predict whether the
website in question is malicious or benign.

\begin{figure}[t]
  \centering
  \includegraphics[width=\columnwidth]{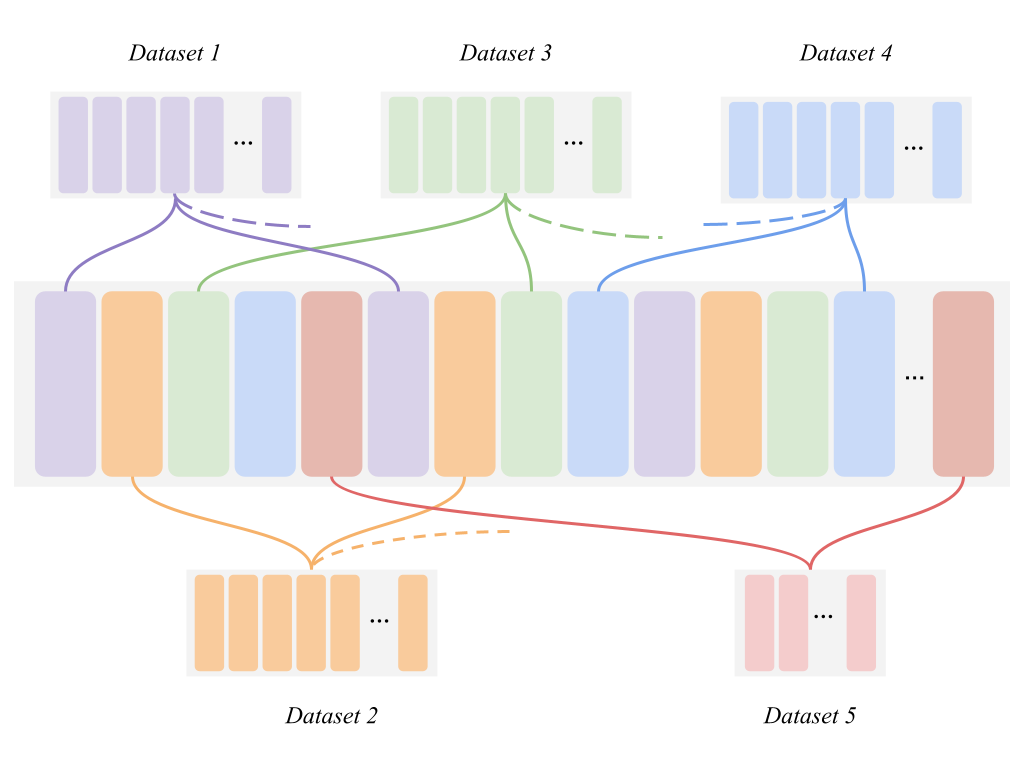}
  \caption{\label{fig:data-split} Divide the data set into the five
    computation nodes. The division must respect the order of the data for
    each node.}
\end{figure}

In order to facilitate our experiment, the dataset was partitioned among the
$5$ and $7$ nodes in the network. As previously mentioned, in some tests the
partition is disjoint but uneven: one of the nodes receives a smaller share of
the data ($50$ samples; the other nodes handle $300$ or $200$ samples each),
so as to determine whether this particular node learns faster when it is
allowed to enrich its observed data with updates and prototypes learnt by its
neighbors. This poses a problem, since incremental datasets are sequential,
and this temporal meaning has to be preserved after the split. We have avoided
this by splitting the stream of data in subsequences, i.e., each node receives
a subsequence of the complete stream, and the samples in each subsequence keep
the same order. The latter means that they are input to ILVQ exactly in the
same temporal order they have originally. This way, semantics of the data is
preserved.  Figure~\ref{fig:data-split} depicts the general procedure for the
data partitioning. Note that round robin (or any other deterministic)
assignment of subsequences to each dataset is not implied.

\section{Results}
\label{sec:results}

This Section presents the results and insights of our simulation
experiments. Specifically, Section~\ref{sec:rsp} presents the performance
achieved under he random sharing protocol (Algorithm~\ref{alg:rsp}) for both
learning models ILVQ and ILVQ+ARF. Subsection~\ref{sec:csp} is analogous: here
we consider the performance of the contextual sharing protocol
(Algorithm~\ref{alg:rtp}) for the two same learning models.

\subsection{XuILVQ memory consumption and runtime}
The ILVQ algorithm was implemented (\textit{XuILVQ}) with the modifications
proposed in Section~\ref{sec:ml_algorithms}. The present study validates this
implementation by comparing its execution time and memory consumption in
different scenarios. Furthermore, the algorithm's performance during
incremental training on synthetic datasets is analyzed, and its results are
compared to those of other classifiers available in River.

We varied the number of features and classes in our generated datasets while
maintaining a binary classification problem in one case and a 10-space dataset
in the other. Our analysis revealed a direct relationship between the memory
usage of the model and the dataset dimension, which may be due to the need to
include more prototypes to accurately represent the dataset in sparse,
high-dimensional spaces. As the number of prototypes increases, the algorithm
spends more time searching for the closest ones, which impacts runtime.

\begin{figure}[t]
  \centering
  \includegraphics[width=0.49\columnwidth]{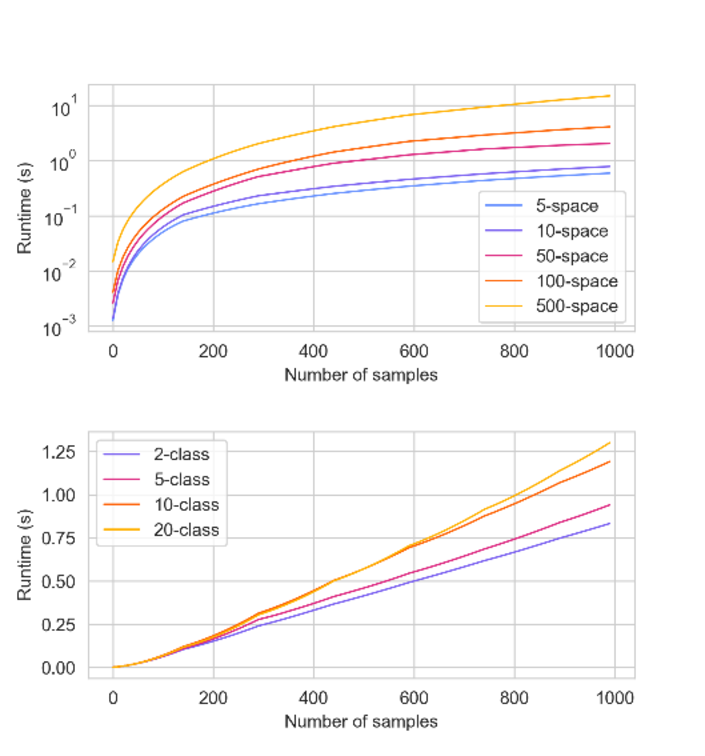}
  \includegraphics[width=0.49\columnwidth]{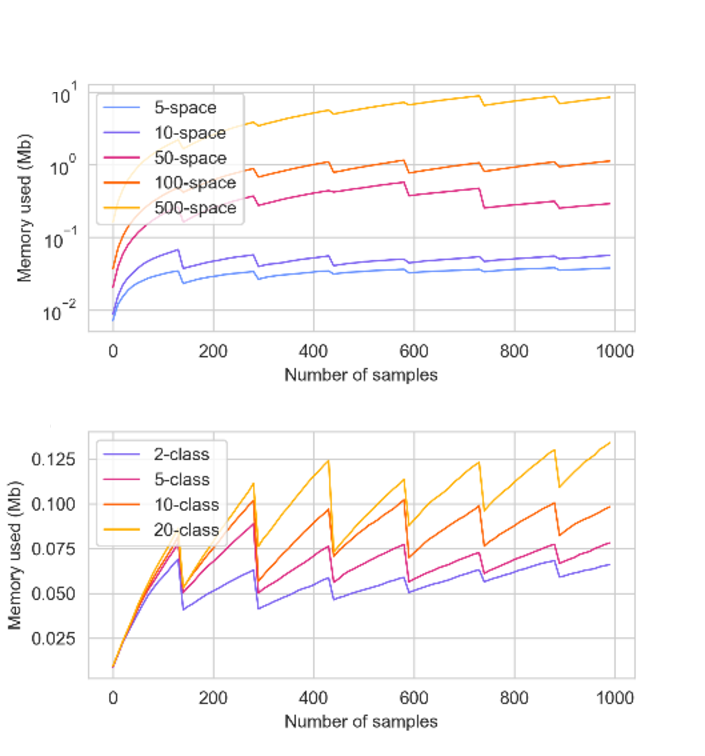}
  \caption{\label{fig:rtp} Evolution of computation time usage as a function
    of the number of features contained in the dataset (top left) and as a
    function of the number of classes contained in the dataset (bottom left)
    for the ILVQ+ARF approach. Evolution of memory consumption as a function
    of the number of features contained in the dataset (top right) and as a
    function of the number of classes in the dataset (bottom right) for the
    ILVQ+ARF approach.}
\end{figure}

Moreover, we observed that memory consumption increases with the number of
classes in the datasets, since the algorithm requires more prototypes to
represent different classes. We also noticed sudden drops in memory usage
forming a sawtooth wave in our results. This is due to the use of the denoise
method, which eliminates obsolete prototypes at a fixed frequency that can be
adjusted by the user (the parameter $\lambda$ in Algoritgm~\ref{alg:ilvq}).

Finally, we compared the memory consumption and runtime of various online
learning algorithms available in River as a function of dataset dimension. All
of the algorithms exhibited linear scaling in both time and memory when
varying the number of variables. Hoeffding Adaptive Tree Classifier (HATC) had
the lowest memory consumption in all scenarios, followed by the XuILVQ
algorithm in most cases (except for high-dimensional scenarios). Regarding
runtime, KNN was faster for lower-dimensional scenarios, while XuILVQ was
faster for high-dimensional ones.

\begin{figure}[t]
  \centering
  \includegraphics[width=0.49\columnwidth]{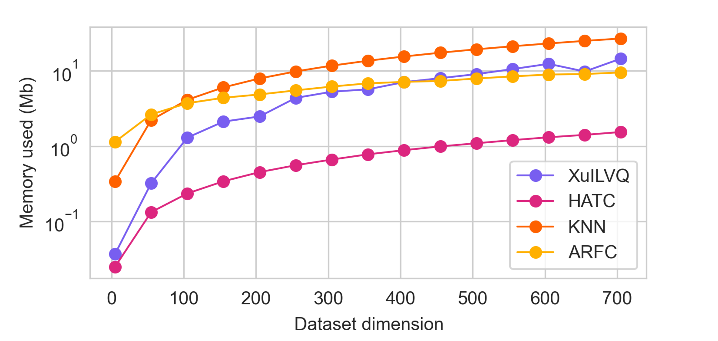}
  \includegraphics[width=0.49\columnwidth]{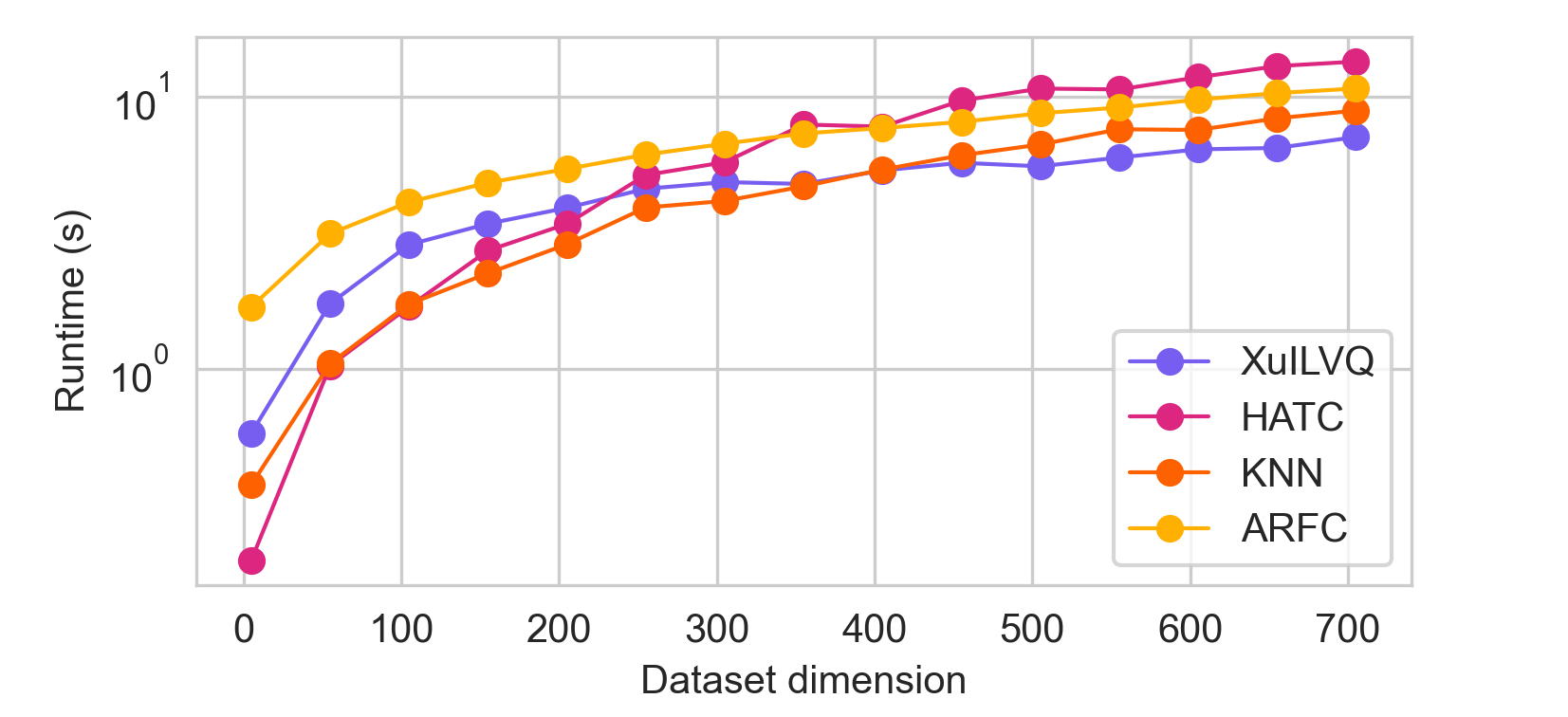}
  \caption{\label{fig:comp} Memory consumption (left) and runtime (right) of
    different incremental learning algorithms compared.  for ILVQ+ARF.}
\end{figure}

\subsection{Random sharing protocol}
\label{sec:rsp}

Following the architecture presented in the preceding section, we configured
five nodes, each one equipped with a model generated either by ILVQ or
ILVQ+ARF. Each node was assigned a dataset that would be used to feed the
algorithm, sample-by-sample. All datasets were of equal size ($300$/$200$
samples), except for one that has only $50$ points. This particular dataset is
chosen as the focus of the study.

Our initial hypothesis is that the node associated with the smallest amount of
data would perform poorly both in terms of accuracy/F-score and in the absence
of data sharing. However, as the sharing parameters of the protocol were
varied, the performance of the model would improve due to the information
encoded in the prototypes identified by its neighbor nodes. To verify this and
to avoid introducing unwanted noise and preventing sharing with the node under
study, the protocol was configured to keep the parameter $s$
unchanged. Therefore, at the epochs when the sharing of prototypes is
executed, the value of $s$ will be broadcast with a frequency determined by
$T$.

\begin{figure}[t]
  \centering
  \includegraphics[width=0.49\columnwidth]{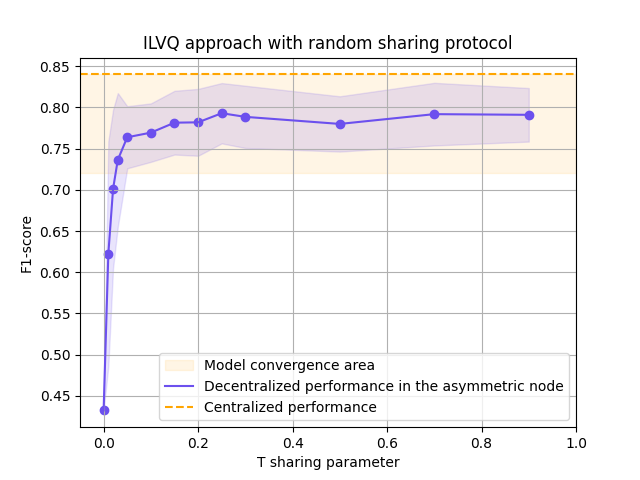}
  \includegraphics[width=0.49\columnwidth]{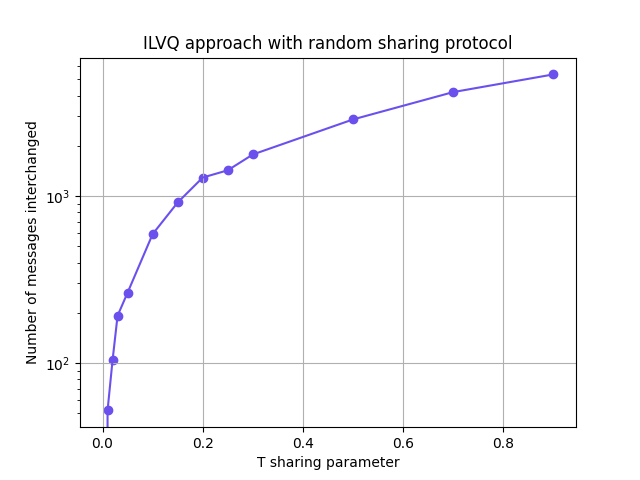}
  \caption{\label{fig:rsp} Random sharing protocol: F1 score (left) and
    communication complexity $L$ for ILVQ with 5 nodes.}
\end{figure}

\begin{figure}[t]
  \centering
  \includegraphics[width=0.49\columnwidth]{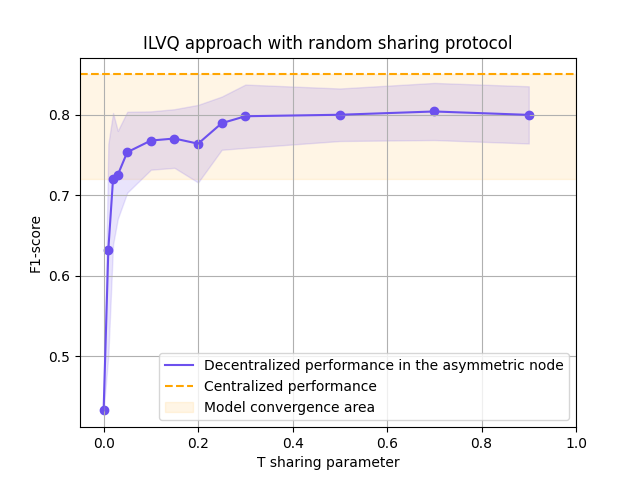}
  \includegraphics[width=0.49\columnwidth]{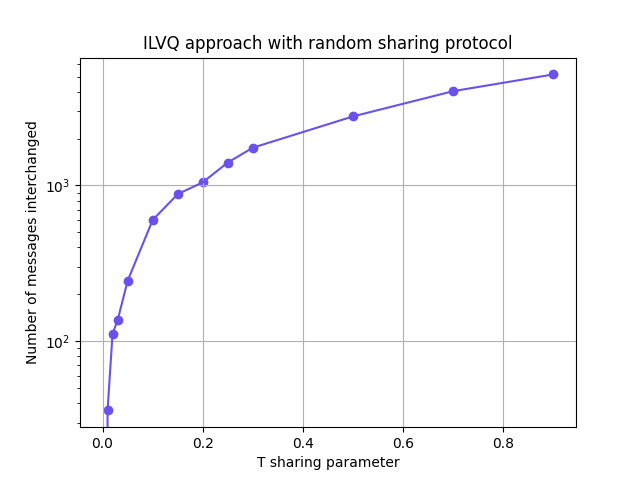}.
  \caption{\label{fig:rsp7} Random sharing protocol: F1 score (left) and
    communication complexity $L$ for ILVQ with 7 nodes.}
\end{figure}

The F1 score results for the asymmetric node were plotted for several values
of $T$ in Figure~\ref{fig:rsp} and ~\ref{fig:rsp7}. In both scenarios, a
fairly similar behavior can be observed. Each point in the graph represents
the average of $20$ simulation runs, providing statistical significance to the
experiment. The results show clearly that, without sharing ($T=0$), the
model's performance is poor, and increases sharply with a moderate amount of
information exchange ($T \approx 0.2$), as expected. Further, the
decentralised operation attains, in the steady state regime ($T \gtrsim 0.3$)
a gap below $10\%$ of the ideal centralised F-score, but using less than $1/5$
of training samples with the latter. Minor fluctuations ($<5\%$) are observed
after the reach of stationary state ($T\approx0.2$)

The figures also shows the growth of the communication complexity $L$ as a
function of $T$. For this small network, $L$ is approximately linear. When
compared with the left panel in Figure~\ref{fig:rsp}, it becomes obvious that
there exists a clear trade-off between communication load and accuracy (more
precisely, F-score). The plots suggest that, since the F-score is
approximately concave in $T$, and $L$ is an affine function of $T$ (roughly),
the F-score is approximately a concave function of $L$.  In turn, this implies
the existence of a maximum point $(L^\ast, F(L^\ast))$ such that further
increases of $T$ (equivalently, $L$) would not yield any significant
improvement in the F-score.

\begin{figure}[t]
  \centering
  \includegraphics[width=0.49\columnwidth]{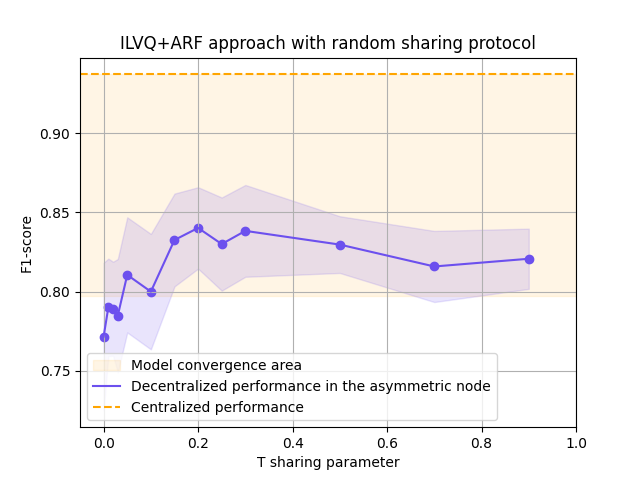}
  \includegraphics[width=0.49\columnwidth]{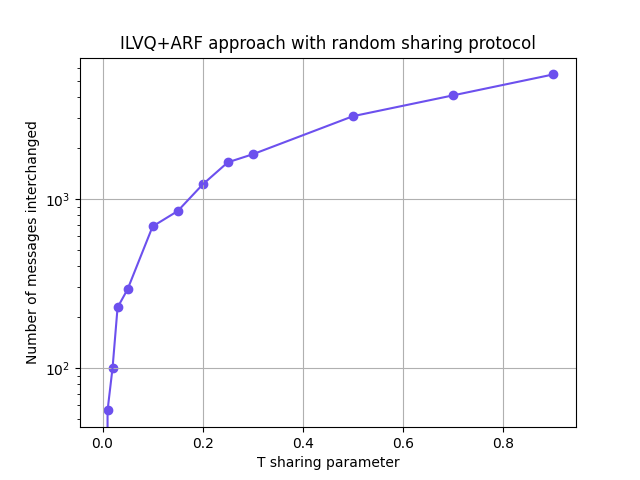}
  \caption{\label{fig:rsp-arf} Random sharing protocol: F1-score (left) and
    communication complexity (right) for ILVQ+ARF with 5 nodes.}
\end{figure}

\begin{figure}[t]
  \centering
  \includegraphics[width=0.49\columnwidth]{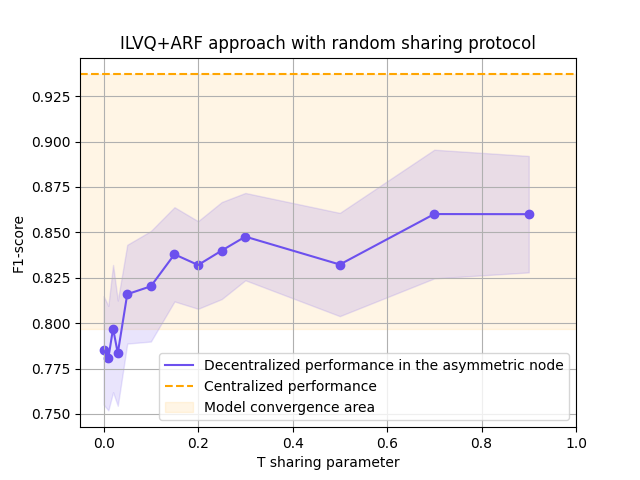}
  \includegraphics[width=0.49\columnwidth]{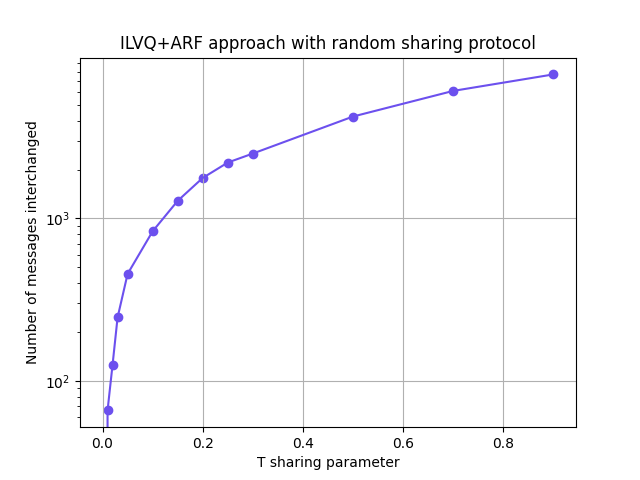}
  \caption{\label{fig:rsp-arf7} Random sharing protocol: F1-score (left) and
    communication complexity (right) for ILVQ+ARF with 7 nodes.}
\end{figure}

Figure~\ref{fig:rsp-arf} and ~\ref{fig:rsp-arf7} depict the results obtained
for the random sharing protocol and the ILVQ+ARF learning algorithm in the
nodes. Similar observations can be drawn here: a moderate value of $T$, the
parameter that controls the degree of sharing, suffices for achieving a
reasonable percentage of the performance with a centralised approach, and the
communication complexity continues to be a linear function of $T$. Actually,
$L$ is the same in both cases, it depends only on the gossip algorithm. The
observed deviations in the ILVQ+ARF configuration from the trend are minimal
and when compared to the confidence interval can be disregarded as
experimental fluctuations.

\subsection{Relative threshold protocol}
\label{sec:csp}

\begin{figure}[t]
  \centering
  \includegraphics[width=0.49\columnwidth]{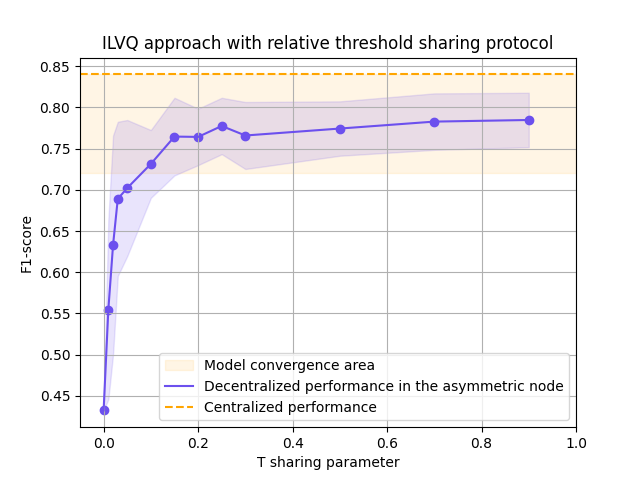}
  \includegraphics[width=0.49\columnwidth]{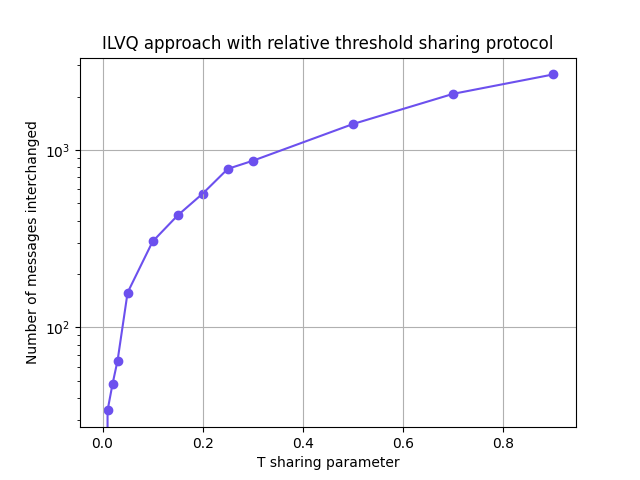}
  \caption{\label{fig:rtp} Relative threshold protocol: F1-score (left) and
    communication complexity (right) for ILVQ with 5 nodes.}
\end{figure}

\begin{figure}[t]
  \centering
  \includegraphics[width=0.49\columnwidth]{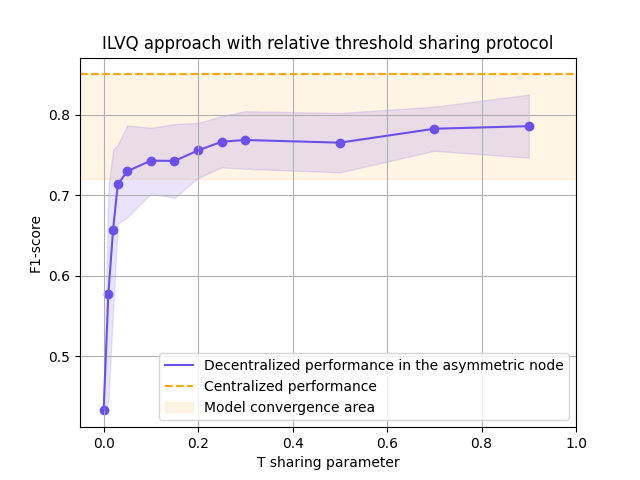}
  \includegraphics[width=0.49\columnwidth]{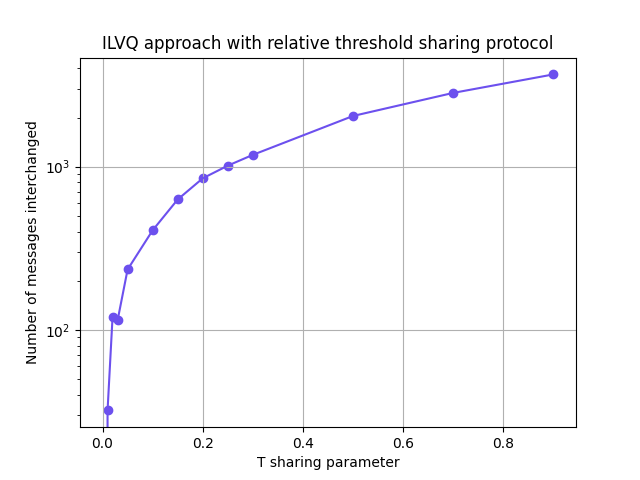}
  \caption{\label{fig:rtp7} Relative threshold protocol: F1-score (left) and
    communication complexity (right) for ILVQ with 7 nodes.}
\end{figure}

\begin{figure}[t]
  \centering
  \includegraphics[width=0.49\columnwidth]{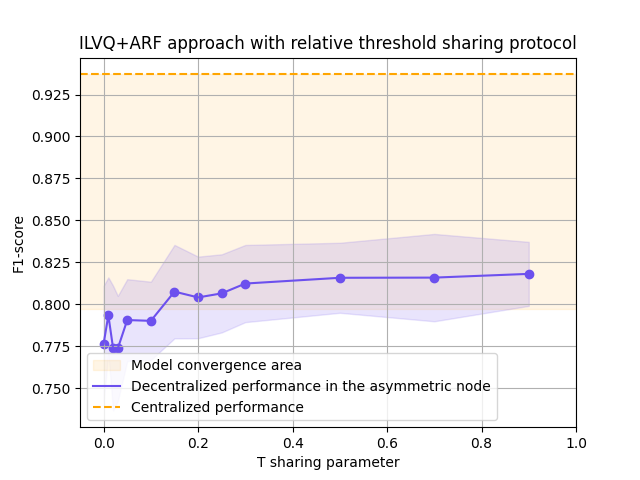}
  \includegraphics[width=0.49\columnwidth]{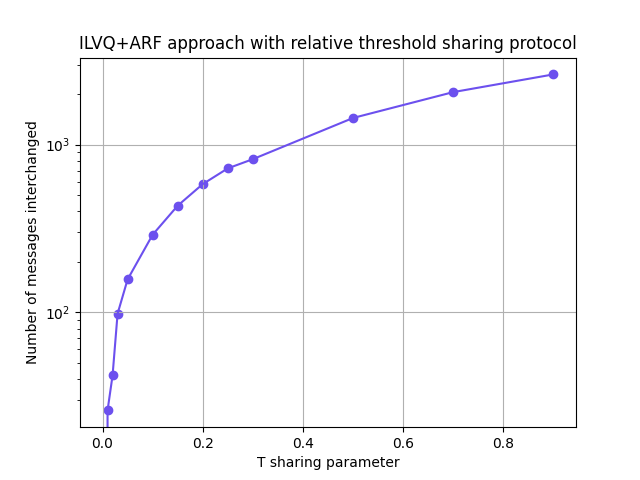}
  \caption{\label{fig:rtp-arf} Relative threshold protocol: F1-score (left)
    and communication complexity (right) for ILVQ+ARF with 5 nodes.}
\end{figure}

\begin{figure}[t]
  \centering
  \includegraphics[width=0.49\columnwidth]{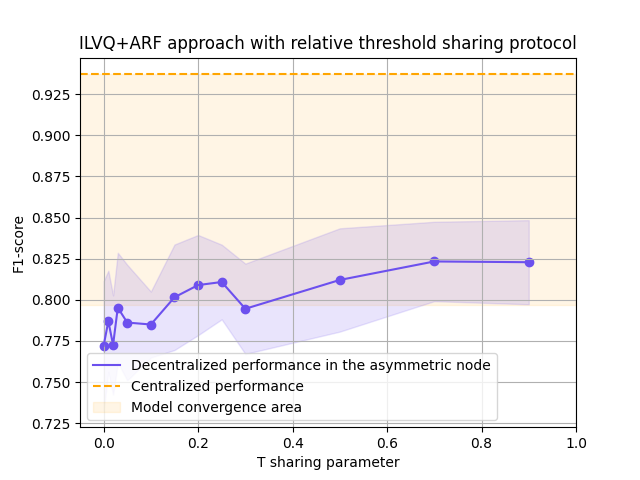} 
  \includegraphics[width=0.49\columnwidth]{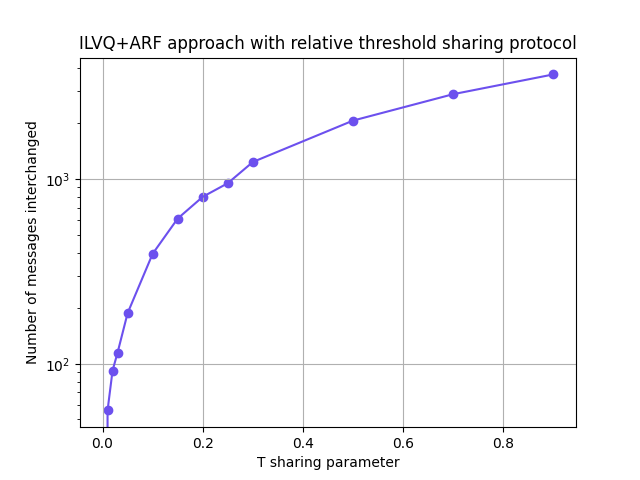}
  \caption{\label{fig:rtp-arf7} Relative threshold protocol: F1-score (left)
    and communication complexity (right) for ILVQ+ARF with 7 nodes.}
\end{figure}

We repeated the experiments after replacing the random sharing by the relative
threshold protocol.  Figures~\ref{fig:rtp} -~\ref{fig:rtp7} and
Figures~\ref{fig:rtp-arf} -~\ref{fig:rtp-arf7} present the results for ILVQ
and ILVQ+ARF, respectively.  The behavior and performance are quite similar to
the random sharing protocol. A possible explanation to this insensitivity is
that the simulated network is small and homogeneous (the tagged node excluded,
of course), so the nodes satisfying the condition of the relative threshold
are not much different from a random sample among all the nodes. This seems to
suggest that a sophisticated gossip or sharing algorithm brings only a
marginal improvement in convergence time and accuracy, and likewise minor
reductions in the communication load. However, we remark that our network is a
complete graph, thus symmetric, so the dynamics of decentralised learning can
be different in more complex topologies.

\section{Conclusions}
\label{sec:conclusions}

This article explores decentralized machine learning and its dynamics through
two meticulously conducted case studies. The investigations concentrate on the
influence of data sharing protocols on machine learning performance. The
random sharing and relative threshold sharing protocols are highlighted,
providing valuable insights into their impacts.

The case studies indicate that the selection of an optimal sharing parameter
plays a critical role in shaping the performance of models within
decentralized machine learning frameworks. Achieving a delicate balance
between maximizing performance gains and preventing network overload is a key
consideration. The studies indicate that a sharing level between 0.05 and 0.2
is the optimal range, providing satisfactory outcomes. Interestingly, moving
beyond this range towards higher sharing parameters results in only minor
enhancements in performance.

This thorough examination highlights the intricate decision-making process
that practitioners must undertake when setting sharing parameters for
decentralized machine learning. The importance of accuracy in this process is
apparent, as it directly impacts the performance of the model and ensures the
efficient use of network resources.As the digital environment continues to
develop, it is essential to refine and optimize protocols for decentralized
machine learning.The insights obtained from these case studies serve as
helpful references for researchers and practitioners who navigate the complex
area of decentralized machine learning.The careful choice of sharing
parameters is essential for achieving peak performance and resource
efficiency, underscoring its critical role in decentralized machine
learning.This study adds to the expanding knowledge base focused on advancing
the field and strengthening its groundwork for future innovations.

\section*{Acknowledgments}
This work was supported by the Axencia Galega de Innovación
(GAIN) (25/IN606D/2021/2612348) and by the Spanish Government
under the research project “Enhancing Communication Protocols
with Machine Learning while Protecting Sensitive Data (COMPROMISE)
(PID2020-113795RBC33/ AEI/10.13039/501100011033).

\appendix

\section{Mathematical justification of protocols}\label{secA1}

Let us assume an epidemic infection model SI that will model rumor-spreading
behavior. In this model there are only two states, susceptible (not informed)
and infected (informed). A susceptible node is a node that has not been
informed but could be informed in the future, this represents those nodes that
have not been exposed to a rumor. An informed node is one that has been
exposed, in our case to the rumor. This simple model is chosen and not another
epidemic model because in our case it is of no interest to represent other
states, nor other state transitions.  The differential equation for the rate
of change of $X$ (informed) and $S$ (not-informed) are:
\begin{equation}
  \frac{dX}{dt} = \beta \frac{SX}{n}, \quad \frac{dS}{dt} = - \beta
  \frac{SX}{n} 
\end{equation}

It is useful to define variables that represent the fractions of not informed
and informed as follows:
\begin{equation}
    x = \frac{X}{n'}, \quad s = \frac{S}{n'}
\end{equation}
\begin{equation}
    \frac{dx}{dt} = \beta sx, \quad \frac{ds}{dt} = - \beta sx
\end{equation}
\begin{equation}
    \frac{dx}{dt} = \beta (1-x)x
\end{equation}
\begin{equation}
    x(t) = \frac{x_0 e^{\beta t}}{1-x_0+x_0 e^{\beta t}}
\end{equation}

For our case study, we assume that $x_0$ is always $1/n'$ because we are
analyzing the spread of information from one node to the rest of the nodes in
the network. Moreover, the probability is not identical in our case since the
probability that the rumor spreads to a new node depends on the value of t
(which is similar to the parameter ) and also depends on the likelihood of
being selected to share the rumor among the node's neighbors who receive the
message. In any case, the model has a late-time property that can be
summarized in that any node within the network that can be informed will end
up being informed regardless of the rumor transmission rate.
\begin{equation}
    x(t)_{t \rightarrow \infty} = \frac{x_0 e^{\beta t}}{1-x_0+x_0 e^{\beta t}} = 1
\end{equation}

When transferring this analysis to a network, the only condition that a
network must have for all nodes within it to be informed is that at least one
vertex must be connected to at least one informed individual by at least one
path through the network, so that the rumor can reach them. Since these are
connected networks, this condition will always be fulfilled and therefore any
rumor transmitted by a node within the network will eventually reach all the
nodes within it. Of course, this oversimplifies the real case in the
simulation since we assume the propagated message is immutable in time, which
is not accurate in our case. The purpose of this analysis is to provide
mathematical foundations for the development of this communication protocol.

\bibliographystyle{elsarticle-num} 
\bibliography{XuILVQ}

\end{document}